P. Wenger,
D. Chablat
M. Zein

Institut de Recherche en Communications et Cybernétique de Nantes

1, rue la Noe – BP 92101 44321 Nantes Cedex 3, France

# Degeneracy study of the forward kinematics of planar 3-RPR parallel manipulators

This paper investigates two situations in which the forward kinematics of planar 3-RPR parallel manipulators degenerates. These situations have not been addressed before. The first degeneracy arises when the three input joint variables $\rho_1$, $\rho_2$ and $\rho_3$ satisfy a certain relationship. This degeneracy yields a double root of the characteristic polynomial in $t = \tan(\varphi/2)$, which could be erroneously interpreted as two coalesce assembly modes. But, unlike what arises in non-degenerate cases, this double root yields two sets of solutions for the position coordinates $(x, y)$ of the platform. In the second situation, we show that the forward kinematics degenerates over the whole joint space if the base and platform triangles are congruent and the platform triangle is rotated by 180 deg about one of its sides. For these "degenerate" manipulators, which are defined here for the first time, the forward kinematics is reduced to the solution of a 3$^{rd}$-degree polynomial and a quadratics in sequence. Such manipulators constitute, in turn, a new family of analytic planar manipulators that would be more suitable for industrial applications.

## 1 Introduction

Solving the forward kinematic problem of a parallel manipulator often leads to complex equations and non analytic solutions, even when considering planar 3-DOF parallel manipulators [1]. For these planar manipulators, Hunt showed that the forward kinematics admits at most 6 solutions [2] and several authors [3, 4] have shown independently that their forward kinematics can be reduced as the solution of a characteristic polynomial of degree 6. In [3], a set of two linear equations in the position coordinates $(x, y)$ of the moving platform is first established, which makes it possible to write $x$ and $y$ as function of the sine and cosine of the orientation angle $\varphi$ of the moving platform. Substituting these expressions of $x$ and $y$ into one of the constraint equations of the manipulator and using the tan-half angle substitution leads to a 6$^{th}$-degree polynomial in $t = \tan(\varphi/2)$. Conditions under which the degree of this characteristic polynomial decreases were investigated in [5, 6]. Four distinct cases were found, namely, (i) manipulators for which two of the joints coincide (ii) manipulators with similar aligned platforms (iii) manipulators with nonsimilar aligned platforms and, (iv) manipulators with similar triangular platforms. For cases (i), (ii) and (iv) the forward kinematics was shown to reduce to the solution of two quadratics in cascade while in case (iii) it was shown to reduce to a 3$^{rd}$-degree polynomial and a quadratic in sequence. To the best of the author's knowledge, no other degenerate cases have been identified yet. In this paper, we show that the forward kinematics of planar 3-RPR[1] manipulators degenerates under conditions that have not been identified before. More precisely, the system of linear equations in $x$ and $y$ that needs be established prior to the derivation of the characteristic polynomial becomes singular under certain conditions. Moreover, we show that the forward kinematics may degenerate over the whole joint space.

Next section recalls the kinematic equations and points out the singularity that may occur when solving the system of linear equations. Section 3 derives the first degeneracy condition, which arises when the input joint coordinates satisfy a certain relationship. The second degeneracy condition is set in section 4. This condition pertains to the geometry of the manipulator and is shown to define new analytic manipulators. These manipulators have a characteristic polynomial of degree 3 instead of 6, and feature more simple singularities. Last section concludes this paper.

## 2 Kinematic equations

Figure 1 shows a general 3-RPR manipulator, constructed by connecting a triangular moving platform to a base with three RPR legs. The actuated joint variables are the three link lengths $\rho_1$, $\rho_2$ and $\rho_3$. The output variables are the position coordinates $(x, y)$ of the operation point $P$ chosen as the attachment point of link 1 to the platform, and the orientation $\varphi$ of the platform. A reference frame is centred at $A_1$ with the $x$–axis passing through $A_2$. Notation used to define the geometric parameters of the manipulator is shown in Fig. 1.

---

[1] The underlined letter refers to the actuated joint

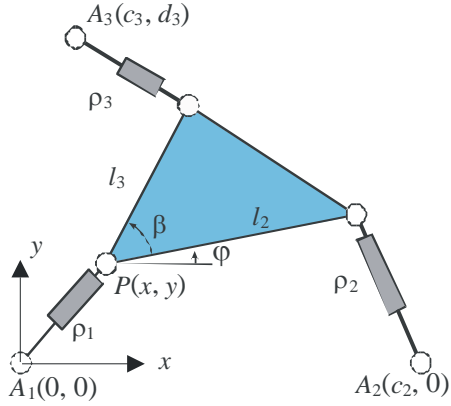

Figure 1: A 3-RPR parallel manipulator

The three constraint equations of the manipulator can be expressed as [3]

$$\rho_1^2 = x^2 + y^2 \tag{1}$$

$$\rho_2^2 = \left(x + l_2 \cos(\varphi) - c_2\right)^2 + \left(y + l_2 \sin(\varphi)\right)^2 \tag{2}$$

$$\rho_3^2 = \left(x + l_3 \cos(\phi + \beta) - c_3\right)^2 + \left(y + l_3 \sin(\phi + \beta) - d_3\right)^2 \tag{3}$$

A system of two linear equations in $x$ and $y$ is first derived by subtracting Eq. (1) from Eqs. (2) and (3), thus obtaining

$$Rx + Sy + Q = 0 \tag{4}$$

$$Ux + Vy + W = 0 \tag{5}$$

where,

$R = 2l_2 \cos(\varphi) - 2c_2$  $\quad U = 2l_3 \cos(\varphi + \beta) - 2c_3$

$S = 2l_2 \sin(\varphi)$  $\quad V = 2l_3 \sin(\varphi + \beta) - 2d_3$

$Q = -2c_2 l_2 \cos(\varphi)$  $\quad W = -2d_3 l_3 \sin(\varphi + \beta) - 2c_3 l_3 \cos(\varphi + \beta)$

$\quad + l_2^2 + c_2^2 - \rho_2^2 + \rho_1^2$  $\quad + l_3^2 + c_3^2 + d_3^2 - \rho_3^2 + \rho_1^2$

As pointed out in [7], $x$ and $y$ can be solved only if the determinant $RV-SU$ is different from zero. If it is so, the 6th-degree characteristic polynomial is obtained upon substituting the expressions of $x$ and $y$ into Eq. (1). The general expression of this characteristic polynomial is not reported here but can be found in [8]. Otherwise, the system degenerates and the forward kinematics cannot be solved this way. To the best of the authors' knowledge, however, the degenerate case $RV-SU = 0$ has never been examined. In the following sections, we will investigate the conditions under which the determinant of the linear system vanishes and will derive the forward kinematics equations associated with these conditions.

## 3 First degeneracy condition

### 3.1 Derivation of the condition

Since $RV-SU$ depends only on the geometric parameters of the manipulator and the orientation $\varphi$ of the platform, $RV-SU = 0$ yields a condition on $\varphi$ for the linear system to degenerate. This condition is

$$c_2 d_3 + l_2 l_3 \sin(\beta) - (l_2 d_3 + c_2 l_3 \sin(\beta))\cos(\varphi) + (l_2 c_3 - c_2 l_3 \cos(\beta))\sin(\varphi) = 0 \tag{6}$$

Resorting to the tan-half angle substitution provides a quadratic in $t = \tan(\varphi/2)$, which has the following form

$$(d_3(l_2 + c_2) + l_3 \sin(\beta)(l_2 + c_2))t^2 + 2(l_2 c_3 - c_2 l_3 \cos(\beta))t - d_3(l_2 - c_2) + l_3 \sin(\beta)(l_2 - c_2) = 0 \tag{7}$$

and may define two orientation angles of the platform.

In order to be able to check the degeneracy condition while solving the forward kinematics, it is useful to set it in terms of $\rho_1$, $\rho_2$ and $\rho_3$. When the determinant of the linear system of Eqs. (4, 5) vanishes, the Cramer's rule tells us that an additional condition must be satisfied for a solution to exist, that is $SW - VQ = 0$, or, equivalently, $RW - UQ = 0$. When set in terms of $t = \tan(\varphi/2)$, this condition yields

$$\left(l_3(2c_2 l_2 + c_2^2 - \rho_2^2 + l_2^2 + \rho_1^2)\sin(\beta) + d_3(2c_2 l_2 + c_2^2 - \rho_2^2 + l_2^2 + \rho_1^2)\right)t^4$$

$$+ \begin{pmatrix} 4l_2 d_3 l_3 \sin(\beta) - 2l_3(l_2^2 + c_2^2 + 2l_2(c_2 - c_3) + \rho_1^2 - \rho_2^2)\cos(\beta) \\ + 2l_2(\rho_1^2 - \rho_3^2 + c_3^2 + d_3^2 + l_3^2) \end{pmatrix} t^3 +$$

$$\left(-4l_2 l_3(c_2 - 2c_3)\sin(\beta) - 8l_2 d_3 l_3 \cos(\beta) + 2d_3(l_2^2 + c_2^2 + \rho_1^1 - \rho_2^2)\right)t^2 +$$

$$\begin{pmatrix} -4l_2 d_3 l_3 \sin(\beta) - 2l_3(l_2^2 + c_2^2 + 2l_2(c_2 - c_3) + \rho_1^2 - \rho_2^2)\cos(\beta) + \\ 2l_2(\rho_1^2 - \rho_3^2 + c_3^2 + d_3^2 + l_3^2) \end{pmatrix} t +$$

$$-l_3(c_2^2 - \rho_2^2 + l_2^2 + \rho_1^2 - 2c_2 l_2)\sin(\beta) + d_3(c_2^2 + l_2^2 - 2c_2 l_2 + \rho_1^2 - \rho_2^2) = 0 \tag{8}$$

Let $t_1$ and $t_2$ define the two solutions of the quadratic defined by Eq. (7). Substituting $t_1$ and $t_2$ into Eq. (8) yields two degeneracy conditions $D_1$ and $D_2$ that depend only on $\rho_1$, $\rho_2$ and $\rho_3$ and on the geometric parameters. Since Eq. (8) is a polynomial of degree 2 in $\rho_1$, $\rho_2$ and $\rho_3$ and since these variables do not appear in Eq. (7), $D_1$ and $D_2$ are also polynomials of degree 2 in $\rho_1$, $\rho_2$ and $\rho_3$. Expressions of the two degeneracy conditions $D_1$ and $D_2$ are quite large when the geometric parameters are left as variables. When rational values are assigned to the geometric parameters of the manipulator, however, $D_1$ and $D_2$ are simple quadratic equations, which can be solved accurately without any difficulties. A particular case with a simple geometric interpretation will be investigated in § 3.3.

### 3.2 Degenerate forward kinematic solutions

If the input variables satisfy one of the two degeneracy conditions $D_1$ and $D_2$, the inverse kinematic problem needs special attention. First of all, the 6th-degree characteristic polynomial $P(t)$ can still be used to calculate all orientation solutions, provided that no division by $RV-SU$ be performed while deriving $P(t)$. Eqs (4, 5) can always be rewritten as $x(RV-SU) = SW-VQ$ and $y(RV-SU) = RW-UQ$. Multiplying both sides of Eq. (1) by $(RV-SU)^2$ and substituting $x^2(RV-SU)^2$ and $y^2(RV-SU)^2$ by $(SW-VQ)^2$ and $(RW-UQ)^2$, respectively, yields $\rho_1^2(RV-SU)^2 = (SW-VQ)^2 + (RW-UQ)^2$, which, after applying the tan-half substitution $t = \tan(\varphi/2)$, yields the 6th-degree characteristic polynomial $P(t)$. If one of the two degeneracy conditions $D_1$ and $D_2$ is satisfied, then $RV-SU = 0$, $SW-VQ = 0$ and $RW-UQ = 0$ and, thus, $P(t)$ vanishes. This means that the degenerate roots defined by

Eq. (7) are also roots of $P(t)$. Moreover, these roots are double roots of $P(t)$ because $RV-SU$, $SW-VQ$ and $RW-UQ$ are squared in $P(t)$. However, unlike what arises in non-degenerate situations, the existence of this double root does not mean that the manipulator admits two coalesce assembly modes. In effect, Eqs. (4,5) cannot be used to calculate $x$ and $y$ since $RV-SU = 0$, and we show below that the double root is associated with two sets of position coordinates $(x, y)$.

To know which equations should be used for the calculation of $x$ and $y$, several cases need be considered. If $R\neq 0$, $x$ can be expressed as function of $y$ from Eq. (4). This expression is then reported into Eq. (1), which yields a quadratics in $y$ and two solutions for $y$. Then each solution is reported in Eq. (4), which is solved for $x$. In this case we have two sets of position coordinates $(x, y)$ with two distinct values of $x$ and two distinct values of $y$. If $R=0$ and $U\neq 0$, the same procedure can be used by using Eq. (5) instead of Eq. (4). If $R=0$, $U=0$ and $S\neq 0$ (resp. $V\neq 0$), $y$ is directly calculated from Eq. (4) (resp. from Eq. (5)). Its solution is then reported in Eq. (1), which gives two values for $x$. In this case we have two sets of position coordinates $(x, y)$ with two distinct values of $x$ but only one for $y$. Finally, if $R=U=S=V=0$, then Eqs. (4,5) show that $Q$ and $W$ must also equal 0, and thus Eq. (2) yields $x^2+y^2=0$, that is, $x=y=0$. Then Eq. (1) implies that $\rho_1=0$. Equating $R$, $U$, $S$, $V$, $Q$ and $W$ to zero implies that $\rho_2=0$ and $\rho_3=0$. This situation is possible only for a manipulator with congruent platform and base triangles.

3.3  A degeneracy condition with geometric interpretation

We investigate now a particular case where $D1$ simplifies and yields a simple geometric interpretation. Assume that $l_2 = c_2$ (the base and platform triangles of the manipulator have the same base length). Then, $t = 0$ is a root of Eq. (7) because the constant term vanishes. Substituting $t = 0$ into Eq. (8) simplifies considerably this equations as only the constant term remains. This constant term can be factored as $(d_3 - l_3\sin(\beta))((c_2 - l_2)^2 + \rho_1^2 - \rho_2^2)$ and when $l_2 = c_2$, this term simplifies and becomes $(d_3 - l_3\sin(\beta))(\rho_1^2 - \rho_2^2)$. Assuming that $l_2 = c_2$ is the only geometric condition, that is, if $d_3 - l_3\sin(\beta) \neq 0$, then the first degeneracy condition, $D1$, is simply $\rho_1^2 - \rho_2^2 = 0$, or $\rho_1 = \rho_2$ if only positive joint values are assumed. The geometric interpretation of $t = 0$, $l_2 = c_2$ and $\rho_1 = \rho_2$ is that the four-bar mechanism defined by disconnecting leg-rod 3 while keeping $\rho_1$, $\rho_2$ constant, is a parallelogram. The coupler curve of the revolute joint centre of the moving platform associated with leg-rod 3 is a circle. For a constant value of $\rho_3$, the tip of leg-rod 3 traces a circle too and its intersection with the one generated by the parallelogram defines the two assembly modes of the 3-RPR manipulator. This situation is illustrated in the numerical example below.

3.4  Numerical example

Let us study the manipulator defined by $c_2 = l_2 = 2$, $c_3 = 1/2$, $d_3 = 1$, $l_3 = 3/2$ and $\beta = \pi/3$. In this case the first degeneracy condition is $\rho_1 = \rho_2$ (we are in the situation described in § 3.3). For joint values that satisfy $\rho_1 = \rho_2$, the polynomial characteristics factors with $t^2$ as the first factor and the following quartic as the second factor:

$$\begin{pmatrix} 16\rho_2^4 - (196 + 32\rho_3^2 + 48\sqrt{3})\rho_2^2 + 16\rho_3^4 + \\ (24 - 48\sqrt{3})\rho_3^2 + 348\sqrt{3} + 805 \end{pmatrix} t^4 +$$

$$\left( (64 - 24\sqrt{3})\rho_2^2 + (48\sqrt{3} - 32)\rho_3^2 - 12\sqrt{3} - 368 \right) t^3 +$$

$$\left( 32\rho_2^4 - (64\rho_3^2 + 96\sqrt{3} + 112)\rho_2^2 - 64\rho_3^4 - 142 + 120\sqrt{3} + 32\rho_3^4 \right) t^2 +$$

$$\left( (64 - 24\sqrt{3})\rho_2^2 + (48\sqrt{3} - 32)\rho_3^2 - 180\sqrt{3} + 304 \right) t +$$

$$16\rho_2^4 + (84 - 48\sqrt{3} - 32\rho_3^2)\rho_2^2 + 16\rho_3^4 + (48\sqrt{3} - 88)\rho_3^2$$

$$-132\sqrt{3} + 229 = 0$$

The first factor $t^2$ gives the degenerate root $t=0$. The position coordinates are calculated as described in §3.2 (note that here $R=0$). For $\rho_1 = \rho_2 = 1$ and $\rho_3 = 7/10$, the quartic admits four real roots. The associated platform orientation angles and positions are given in Tab. (1), along with the two platform positions associated with the degenerate root $t = 0$. Figure 2 shows the associated 6 assembly-modes. One can easily verify that the two assembly modes associated with the degenerate root $t = 0$ are distinct.

|   | Quartic root #1 | Quartic root #2 | Quartic root #3 | Quartic root #4 | Degenerate root t=0 | |
|---|---|---|---|---|---|---|
| $\varphi$ | -43.8049 deg | -6.6271 deg | 23.6384 deg | 58.4876 deg | 0 | |
| $x$ | -0.3395 | -0.9849 | 0.9768 | 0.6632 | -0.1394 | -0.9499 |
| $y$ | 0.9406 | 0.1728 | -0.2141 | -0.7485 | -0.9902 | -0.3126 |

Table 1: The six sets of solutions for a manipulator defined by $c_2=l_2=2$, $c_3=1/2$, $d_3=1$, $l_3=3/2$, $\beta=\pi/3$ with inputs $\rho_1=\rho_2=1$ and $\rho_3=3/5$.

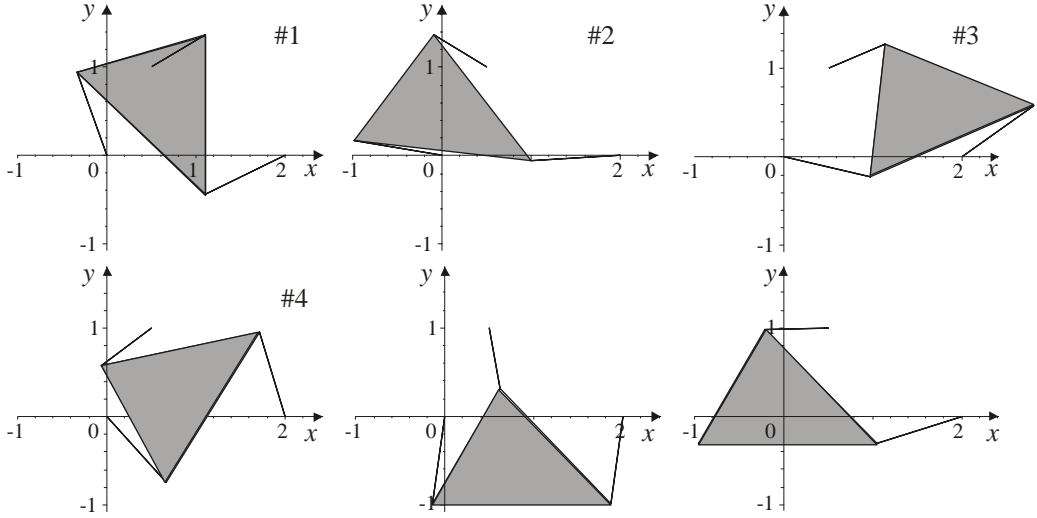

**Figure 2**: The six assembly-modes corresponding to Tab. 1. The last two ones correspond to the degenerate root.

## 4 Degeneracy over the whole joint space

### 4.1 Condition on the manipulator geometry

We would like to know if it is possible to find manipulators for which the system of linear equations (4,5) degenerates for any input joint values. For such manipulators, Eq. (7) must be satisfied for any value of $t$. Thus, the following three conditions must be simultaneously satisfied

$$d_3(l_2 + c_2) + l_3 \sin(\beta)(l_2 + c_2) = 0 \qquad (9)$$

$$l_2 c_3 - c_2 l_3 \cos(\beta) = 0 \qquad (10)$$

$$-d_3(l_2 - c_2) + l_3 \sin(\beta)(l_2 - c_2) = 0 \qquad (11)$$

Eqs. (9) and (11) yield $l_2 = c_2$ and $\sin(\beta) = -d_3/l_3$. Substituting $l_2 = c_2$ into Eq. (10) yields $\cos(\beta) = c_3/l_3$. Thus, one must have $l_3 = \sqrt{d_3^2 + c_3^2}$. It can be easily verified that the geometric interpretation of these conditions is that the base and the platform triangles are congruent and the platform triangle is rotated by 180 deg about the side $l_2$. Note that if we had had $\sin(\beta) = d_3/l_3$ instead of $\sin(\beta) = -d_3/l_3$, we would have got manipulators with congruent base and platform triangles that exhibit a passive translational motion when $\varphi = 0$ [5, 6].

### 4.2 Forward kinematics

Since the system of linear equations is always singular, this means that $RV - SU = 0$ independently of $\rho_1$, $\rho_2$ and $\rho_3$. Thus, $SW - VQ$ must be equal to zero for any value of $\rho_1$, $\rho_2$ and $\rho_3$. This means that $t$ must satisfy Eq. (8). When substituting $l_2 = c_2$, $\sin(\beta) = -d_3/l_3$, $\cos(\beta) = c_3/l_3$ and $l_3 = \sqrt{d_3^2 + c_3^2}$ into Eq. (8), the coefficient of highest degree vanishes and we get the following 3$^{rd}$-degree polynomial in $t = \tan(\varphi/2)$:

$$(c_3(\rho_1^2 - \rho_2^2 + 4c_2^2 - 4c_3 c_2) + c_2(\rho_3^2 - \rho_1^2))t^3 +$$
$$d_3(8c_3 c_2 - 4c_2^2 + \rho_2^2 - \rho_1^2)t^2 + \qquad (12)$$
$$(c_3(\rho_1^2 - \rho_2^2) + \rho_3^2 c_2 - 4d_3^2 c_2 - \rho_1^2 c_2)t + d_3(\rho_2^2 - \rho_1^2) = 0$$

which is the characteristic polynomial to be solved for this family of manipulators. Thus, there exists a new family of analytic manipulators, namely, those that have congruent base and platform triangles with the platform triangle rotated by 180 deg about the side $l_2$. The position coordinates of the platform are then calculated by solving in cascade a quadratic and a linear equation as explained in §3.2. These analytic manipulators may have up to 6 assembly-modes but, unlike general 3-RPR manipulators, they have only three distinct platform orientations and each platform orientation is associated with two distinct positions. Because their forward kinematics degenerates over the whole joint space, we call these manipulators *degenerate manipulators*.

### 4.3 Numerical example

A numerical example is now presented to show the different assembly modes of a given degenerate manipulator. The geometric parameters are $c_2 = l_2 = 1$, $c_3 = 0$, $d_3 = 1$, $l_3 = 1$ and $\beta = -\pi/2$. These parameters satisfy the geometric conditions for the manipulator to be degenerate.

The direct kinematics is now calculated for $\rho_3 = 4/5$, $\rho_2 = \rho_3 = 3/2$. The 3$^{rd}$-degree polynomial is:

$$161t^3 - 239t^2 - 239t + 161 = 0 \qquad (13)$$

The six sets of solutions are reported in Tab. 2.

|  | #1 | #2 | #3 | #4 | #5 | #6 |
|---|---|---|---|---|---|---|
| $\varphi$ (deg) | -90 | -90 | 53.6102 | 53.610 | 126.389 | 126.389 |
| $x$ | 0.6547 | -0.459 | 0.3963 | -0.794 | 0.6950 | 0.0933 |
| $y$ | -0.4597 | 0.6547 | 0.6950 | 0.0933 | 0.3963 | -0.7945 |

**Table 2: The six sets of solutions for a degenerate manipulator defined by $c_2=l_2=1$, $c_3=0$, $d_3=1$, $l_3=1$ and $\beta=-\pi/2$ with inputs $\rho_1=4/5$ and $\rho_2=\rho_3=3/2$.**

## 5   Conclusions

In this paper, the general $6^{th}$-degree characteristic polynomial of planar 3-RPR parallel manipulators was shown to degenerate in situations that had not been investigated in the past.

The first degeneracy may occur for any manipulator geometry. There may exist two platform angles for which the system of linear equations in $x$ and $y$ that needs be established prior to the derivation of the characteristic polynomial, becomes singular. Each of these two angles defines a relationship between the input joint values. When solving the forward kinematics, if the input joint values satisfy one of these two relationships, the general $6^{th}$-degree characteristic polynomial gives a double root, which is a degenerate root for which the aforementioned system of linear equations is singular, and which could be erroneously interpreted as two coalesce assembly modes and, thus, as a kinematic singularity. But one gets an unusual situation in which there is one platform orientation associated with two distinct sets of positions (x, y). The second degenerate situation has an interesting practical consequence. It occurs for any input joint value when the manipulator has a special geometry, namely, congruent base and platform triangles with a rotation of 180 deg of the platform triangle about one of its sides. The forward kinematics of such manipulators can be solved with a $3^{rd}$-degree characteristic polynomial and a quadratic in sequence. Thus, these manipulators, which we have called *degenerate manipulators*, define a new family of analytic manipulators that would be more suitable for industrial applications.

These degenerate manipulators are now being examined in more details by the authors of this paper to evaluate their potential interests for industrial applications. First results show that their singularity locus is much simpler than the one of general manipulators. In addition, contrary to the analytic manipulators with similar platform and base triangles studied in [5, 9] that have also a simple singularity locus, the degenerate manipulators defined here are not singular at $\varphi = 0$ or $\varphi = \pi$ throughout the plane and we have observed that they can execute non-singular assembly-mode changing motions [8, 10].

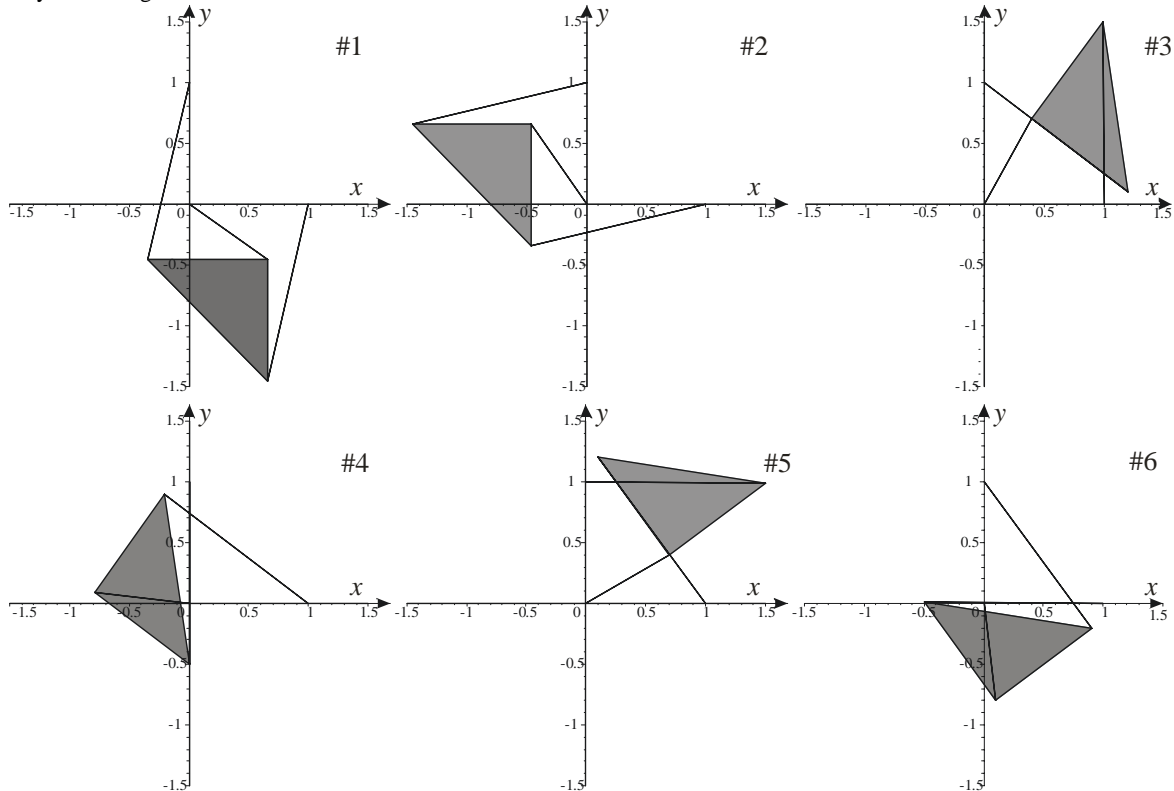

**Figure 3: The six assembly-modes corresponding to Tab. 2.**